\documentclass[10pt,twocolumn,letterpaper]{article}

\usepackage{cvpr}              

\usepackage{graphicx}
\usepackage{amsmath}
\usepackage{amssymb}
\usepackage{booktabs}
\usepackage{multirow}
\usepackage{bm}
\usepackage{makecell}

\usepackage[pagebackref,breaklinks,colorlinks]{hyperref}

\usepackage[capitalize]{cleveref}
\crefname{section}{Sec.}{Secs.}
\Crefname{section}{Section}{Sections}
\Crefname{table}{Table}{Tables}
\crefname{table}{Tab.}{Tabs.}

\usepackage[marginal]{footmisc}

\def\confName{CVPR}
\def\confYear{2023}

\begin{document}
\title{Bridging Severe Cross-Modal Misalignment: End-to-End Visible-Infrared Object Detection via Explicit Feature-Domain Affine Registration}

\author{Qi Ming\textsuperscript{1}, Yuyang Wang\textsuperscript{2}$^\ast$, Mingjing Zhao\textsuperscript{3}, Yifan Xiao\textsuperscript{4}, \\ Zhixin Guo\textsuperscript{4}, Zhiqiang Zhou\textsuperscript{5}, Peng Sun\textsuperscript{6}, Juan Fang\textsuperscript{1}, Fuqiang Yang\textsuperscript{7}, Xudong Zhao\textsuperscript{5}\\
{\small
	\begin{tabular}{ll}
		\textsuperscript{1}Beijing University of Technology, China &
		\textsuperscript{2}Central South University, China \\
		\textsuperscript{3}Beijing Electronics Science \& Technology Institute &
		\textsuperscript{4}China Aerospace Science \& Industry Corporation \\
		\textsuperscript{5}Beijing Institute of Technology, China &
		\textsuperscript{6}Information Support Force Engineering University, China \\
		\multicolumn{2}{c}{
			\textsuperscript{7}Trunk Technology (Beijing) Co., Ltd., China
		}
	\end{tabular}
}\\
{\tt\small chaser.ming@gmail.com, killlakill2023@gmail.com}\\ 
}
\maketitle
\footnote{$^\ast$ Corresponding author.\\}
\begin{abstract}
Visible-infrared object detection relies on complementary RGB and thermal cues, but its performance is often degraded by cross-modal spatial misalignment. Most existing methods rely on implicit feature adaptation to handle weakly misaligned scenarios, while large-offset geometric discrepancies remain insufficiently addressed. In this paper, we propose a Joint Feature-domain Registration and Detection network (JFRDet), an end-to-end visible-infrared oriented object detector tailored for severely cross-modal geometric discrepancies. JFRDet introduces a Cross-Modal Affine Alignment (CMAA) module to estimate an image-level affine transformation for explicit multi-level feature alignment. Note that illumination changes directly affect the reliability of RGB cues, an Illumination-Guided Complementary Fusion (IGCF) module adaptively exploits modality reliability under varying illumination conditions for cross-modal fusion. Then, an Alignment Quality-Consistency Gating (AQCG) strategy stabilizes joint optimization by modulating detection supervision according to alignment reliability and gradient consistency. We further construct DroneVehicle Misaligned (DVMA), a benchmark for evaluating visible-infrared oriented object detection under severe cross-modal geometric misalignment. The proposed JFRDet achieves 69.7\% $\mathrm{mAP}_{50}$ on DVMA, which represents state-of-the-art (SOTA) performance. The code and dataset will be available on GitHub.
\end{abstract}

\section{Introduction}
\label{sec:intro}

Visible-Infrared Object Detection (VIOD) has attracted increasing attention by exploiting the complementary strengths of visible and infrared modalities \cite{hwang2015multispectral, liu2016multispectral}.  RGB images provide rich texture and appearance cues, yet they often encounter limitations under low-light or adverse weather conditions. In contrast, infrared images are more robust to illumination variations, making their combination particularly effective for applications such as surveillance and autonomous driving \cite{jia2021llvip, li2019illumination}. However, visible-infrared image pairs often exhibit spatial misalignment due to differences in sensor placement, viewpoint, or platform motion. Such cross-modal spatial misalignment, including translation, rotation, and scale variations, significantly undermines the effectiveness of cross-modal feature fusion \cite{Zhang_2019_ICCV}. Robust cross-modal fusion under spatial misalignment remains a major challenge.

To address this issue, recent visible-infrared detectors incorporate alignment-aware designs into the detection pipeline. Some methods \cite{Zhang_2019_ICCV,zhang2021weakly,yuan2022translation} estimate cross-modal correspondences to recalibrate modality-specific features, thereby reducing fusion errors caused by misalignment. Other studies \cite{he2023misaligned,song2024misaligned,yuan2024improving} combine explicit calibration with feature correction or region-level reasoning to further alleviate modality discrepancies.
Another line of work avoids strict geometric correction and instead develops fusion modules robust to weak misalignment. Representative strategies include deformable attention and offset-guided sampling \cite{Chen_2024_CVPR, fu2024cf, guo2024damsdet}, which allow the model to adaptively aggregate cross-modal features without precise spatial correspondence.

\begin{figure*}[t] 
	\centering
	\includegraphics[width=1\linewidth]{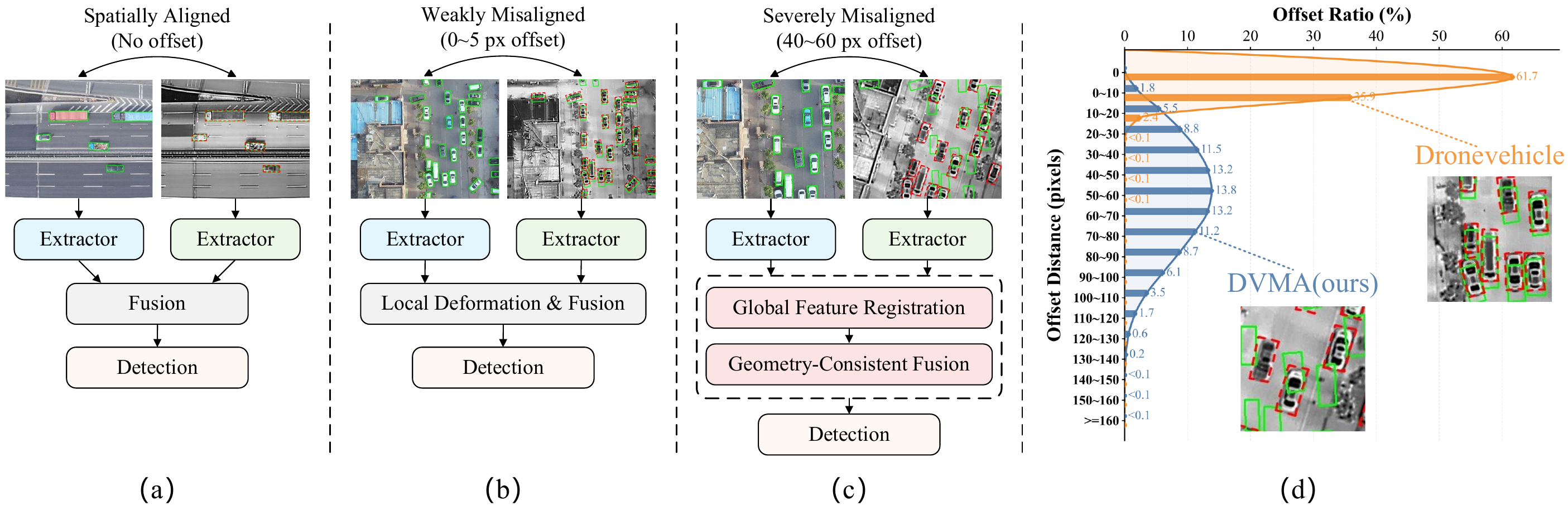} 
	
	\caption{
		Comparison of different multimodal detection paradigms. 
		(a) Spatially aligned multimodal detection assumes well-registered inputs. 
		(b) Weakly misaligned detection performs local pixel- or feature-level alignment before fusion. 
		(c) Our JFRDet predicts an image-level affine transform and progressively applies it to multi-level features, enabling aligned fusion and detection under severe misalignment.
		(d) Offset distribution comparison between DroneVehicle and DVMA.
	}
	
	\label{tasks Comparison} 
\end{figure*}

However, existing misalignment-aware VIOD studies suffer from two key limitations. (1) \textbf{Severe cross-modal misalignment remains underexplored.} Most methods are designed and evaluated on aligned or weakly misaligned image pairs. As shown in Fig \ref{tasks Comparison} (d), most offsets in DroneVehicle\cite{sun2022drone} fall within 1–5 pixels, offering limited evidence of robustness to severe misalignment. (2) \textbf{Feature downsampling obscures weak misalignment.} Repeated backbone downsampling maps small image-level offsets to barely distinguishable displacements on low-resolution feature maps, effectively creating near-aligned training conditions. Consequently, existing methods may not learn to resolve intrinsic cross-modal spatial discrepancies, limiting their robustness in severely misaligned real-world scenes.

To this end, this paper proposes the Joint Feature-domain Registration and Detection Network (JFRDet), an end-to-end framework for VIOD in severe misalignment scenarios. JFRDet enables registration and oriented object detection to be optimized within a unified framework. Specifically, a Cross-Modal Affine Alignment module (CMAA) is introduced to recover the global geometric relationship between modalities by establishing cross-modal correspondences and estimating an affine transformation. Then, an Illumination-Guided Complementary Fusion module (IGCF) adaptively balances visible and infrared cues according to illumination, compensating degraded RGB evidence with infrared responses while retaining useful appearance details. Next, an Alignment Quality-Consistency Gating strategy (AQCG) stabilizes joint optimization by regulating detector learning based on alignment reliability and task consistency. These components jointly enable robust visible-infrared oriented object detection under large cross-modal geometric variations. Moreover, we construct the DroneVehicle Misaligned (DVMA) benchmark to evaluate visible-infrared oriented object detection under severe cross-modal misalignment. Compared with weakly misaligned datasets, DVMA provides a more challenging setting for assessing cross-modal alignment and detection robustness, as illustrated in Fig.~\ref{tasks Comparison} (d).

Our main contributions can be summarized as follows:

1) This paper proposes JFRDet, an end-to-end framework that performs explicit feature-domain affine registration for accurate VIOD. JFRDet is a pioneering work to address visible-infrared oriented object detection under severe cross-modal misalignment scenarios.

2) We introduce JFRDet with CMAA for explicit affine feature registration. IGCF then performs illumination-adaptive cross-modal fusion, while AQCG coordinates registration and detection training by adjusting detection supervision according to alignment quality and optimization consistency.

3) DVMA benchmark is constructed with substantially larger cross-modal geometric discrepancies than existing visible-infrared datasets. It provides a challenging evaluation setting for assessing detection robustness under severe visible-infrared misalignment.

\section{Related Work}
\subsection{Oriented Object Detection}
Oriented object detection is widely studied in aerial remote sensing, objects exhibit arbitrary orientations and dense layouts.Early methods adapt horizontal detectors using rotation-aware proposals or RoI transformations~\cite{ma2018arbitrary,jiang2017r2cnn,ding2019learning}. Subsequent studies improve localization through progressive rotated-box refinement and feature alignment~\cite{yang2021r3det,xie2021oriented}. Others introduce flexible object representations, such as vertex-based or point-set-based representations, to describe arbitrary oriented targets~\cite{xu2020gliding,li2022oriented}. Feature-adaptive strategies are also explored to improve the discrimination of densely arranged objects~\cite{guo2021beyond}. Another line of work reformulates angle prediction or regression losses to mitigate boundary discontinuity and improve localization consistency~\cite{yang2020arbitrary,yang2021dense,yang2021rethinking,yang2021learning,yang2022kfiou}. Despite these advances, most oriented detectors are designed for visible imagery, which limits their robustness under low-light conditions.

\subsection{Visible-Infrared Object Detection}
Visible-infrared object detection exploits the complementary properties of RGB and IR imagery. Early methods mainly focused on where and how to fuse the two modalities, ranging from deep feature fusion and multi-layer fusion to joint detection-segmentation learning \cite{wagner2016multispectral, chen2018multi, li2018multispectral}. Later works further introduced gated fusion, modality balancing, and dynamic cross-modal interaction to suppress unreliable modality responses and enhance complementary cues \cite{zheng2019gfd, zhou2020improving, xie2022learning}. Recently, attention- and Transformer-based designs model long-range cross-modal dependencies and improve interaction in complex scenes~\cite{qingyun2021cross,qingyun2022cross,shen2024icafusion,zhang2024tfdet}. Other studies reduce redundant information or estimate modality confidence during aggregation~\cite{zhao2025removal,li2023multiscale}. Despite improved robustness under low-light conditions, most methods assume well-registered or weakly misaligned image pairs and remain insufficiently robust to severe cross-modal misalignment.

\section{Method}

\begin{figure}[t] 
	\centering
	\includegraphics[width=1\linewidth]{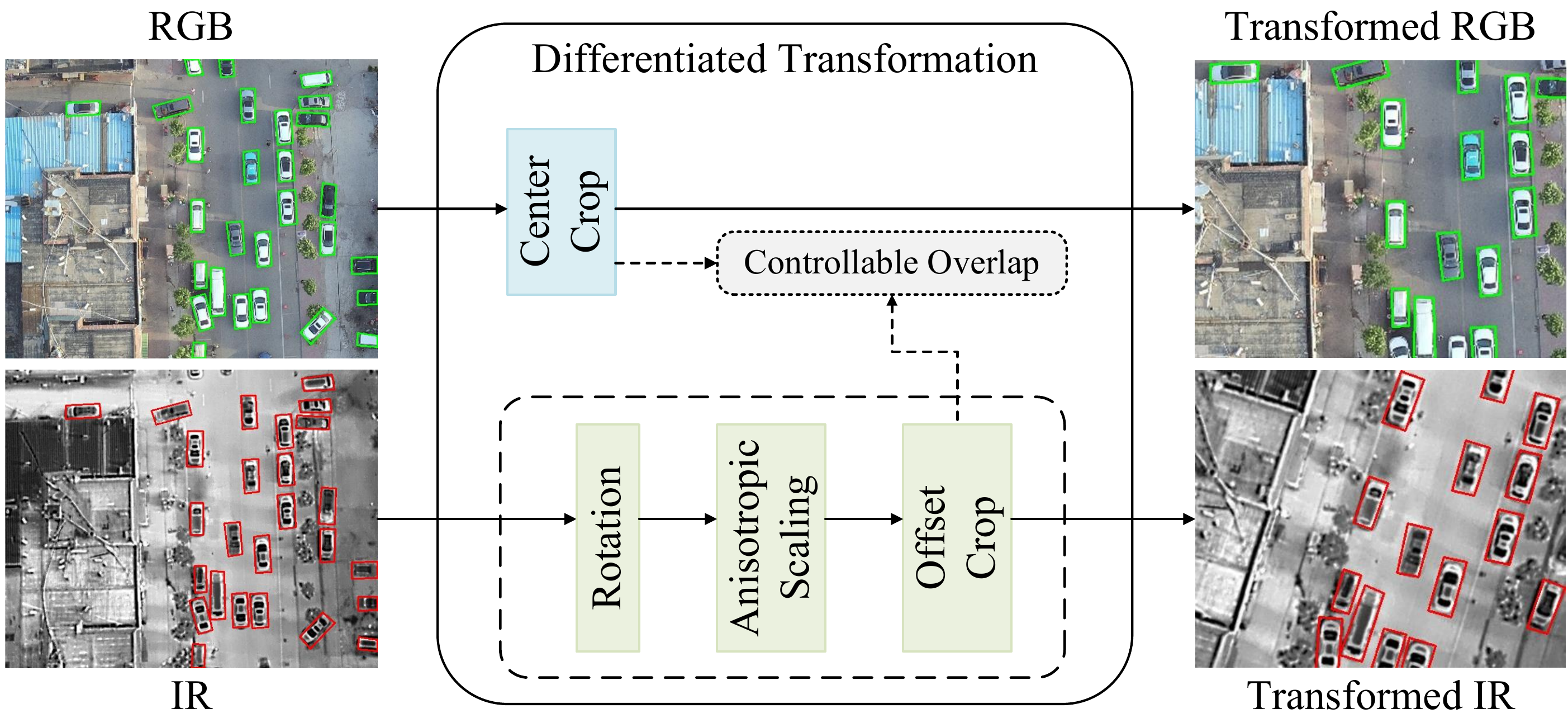} 
	
	\caption{
		Construction pipeline of the DVMA benchmark.
	}
	
	\label{DVMA} 
\end{figure}

\subsection{DroneVehicle Misaligned Benchmark}

As illustrated in Fig.~\ref{DVMA}, DVMA is constructed from aligned image pairs in DroneVehicle~\cite{sun2022drone}. The visible image is center-cropped, while the infrared image undergoes a compound transformation composed of rotation, anisotropic scaling, and translation. The rotation angle is sampled between $15^\circ$ and $30^\circ$ in either direction, and the horizontal and vertical scaling factors are independently sampled from $[0.90,1.10]$. We retain only pairs whose transformed valid regions have an overlap IoU within $[0.70,0.80]$. These transformations produce spatially varying cross-modal offsets ranging from tens to hundreds of pixels while maintaining sufficient scene overlap. Since all transformation parameters are recorded, exact cross-modal correspondences can be generated automatically at each feature resolution. Continuous correspondence coordinates are used to construct normalized ground-truth offset fields consistent with the $[-1,1]$ coordinate range of bilinear sampling, while discretized target locations provide matching labels. Positions mapped outside the valid feature region are masked during loss computation, and correspondence supervision is generated symmetrically in both visible-to-infrared and infrared-to-visible directions. As summarized in Table~\ref{tab:dataset_comparison}, compared with existing misaligned datasets, DVMA provides explicit alignment annotations for paired images, enabling supervised cross-modal registration during training.


\begin{table}[t]
	\centering
	{\scriptsize
		\setlength{\tabcolsep}{2.8pt}
		\begin{tabular}{@{}lccccc@{}}
			\toprule
			\textbf{Dataset}
			& \textbf{Images}
			& \textbf{Misaligned}
			& \textbf{Categories}
			& \textbf{Scenario}
			& \textbf{Alignment GT} \\
			\midrule
			VEDAI
			& 1200
			& $\times$
			& 9
			& drone
			& $\times$ \\
			
			KAIST
			& 95328
			& $\triangle$
			& 3
			& driving
			& $\times$ \\
			
			CVC14
			& 17036
			& $\checkmark$
			& 1
			& driving
			& $\times$ \\

			FLIR
			& 24680
			& $\checkmark$
			& 4
			& driving
			& $\times$ \\
			
			LLVIP
			& 30976
			& $\times$
			& 1
			& surveillance
			& $\times$ \\
			
			DroneVehicle
			& 56878
			& $\triangle$
			& 5
			& drone
			& $\times$ \\
			
			M$^{3}$FD
			& 8400
			& $\times$
			& 6
			& various
			& $\times$ \\
			
			DVTOD
			& 4358
			& $\checkmark$
			& 3
			& drone
			& $\times$ \\
			
			\midrule
			
			\textbf{DVMA (ours)}
			& \textbf{56878}
			& $\checkmark$
			& \textbf{5}
			& \textbf{drone}
			& $\checkmark$ \\
			\bottomrule
		\end{tabular}
	}
	\caption{Comparison of multispectral object detection datasets.      $\triangle$ indicates that only a subset of objects are weakly misaligned. `Alignment GT' indicates whether ground-truth cross-modal correspondences are provided for alignment.}
	\label{tab:dataset_comparison}
\end{table}

\subsection{Overview Architecture}

\begin{figure*}[t] 
	\centering
	\includegraphics[width=1.00\linewidth]{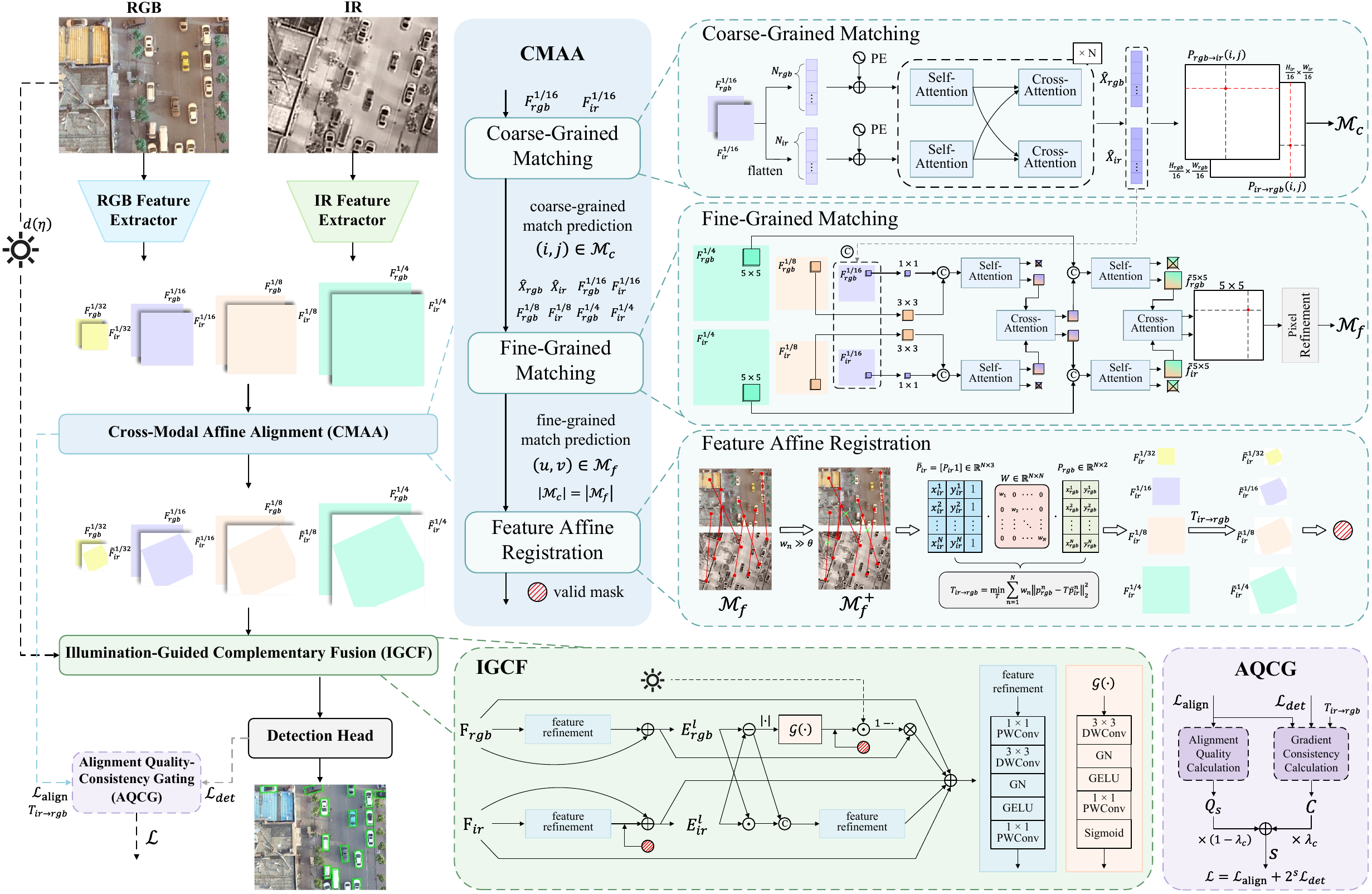} 
	
	\caption{
		Network architecture of the proposed JFRDet.
	}
	
	\label{main} 
\end{figure*}

The overall framework of JFRDet is illustrated in Fig.~\ref{main}. Paired RGB-IR images are processed by a dual-stream backbone following \cite{Zhou2025DMM} to extract multi-scale modality-specific features. Taking the visible branch as reference, CMAA estimates an affine transformation and progressively aligns infrared features. IGCF then fuses the aligned features under varying illumination conditions, which are fed into an S$^{2}$A-Net head~\cite{han2021align} for oriented prediction. To stabilize training, AQCG regulates detection supervision based on alignment reliability, reducing the negative impact of unreliable alignment on detector optimization.

\subsection{Cross-Modal Affine Alignment}

Given multi-scale visible and infrared features ${F_{rgb}^{s}}$ and ${F_{ir}^{s}}$, where $s\in{1/4,1/8,1/16,1/32}$ denotes the resolution relative to the input, the proposed CMAA establishes coarse-to-fine cross-modal correspondences using the $1/4$, $1/8$, and $1/16$ features. Specifically, it first identifies robust coarse correspondences $\mathcal{M}_{c}$ under severe spatial offsets, and then refines them into more precise fine-grained correspondences $\mathcal{M}_{f}$. These correspondences are used to estimate an image-level affine transformation $\mathbf{T}_{ir\rightarrow rgb}\in\mathbb{R}^{2\times3}$, which progressively warps all infrared pyramid features to obtain aligned representations $\tilde{F}_{ir}={\tilde{F}_{ir}^{s}}$. Consequently, subsequent cross-modal fusion is conducted on geometrically aligned features rather than directly on misaligned RGB-IR representations.

\subsubsection{Coarse-Grained Matching}
To establish robust initial correspondences under severe cross-modal spatial offsets, coarse-grained matching is performed on $F_{rgb}^{1/16}$ and $F_{ir}^{1/16}$. After positional encoding and flattening, the resulting tokens are processed by alternating intra-modal self-attention and inter-modal cross-attention to capture long-range context and cross-modal correspondences. The refined sequences are denoted as $\hat{X}_{rgb}=\{\hat{x}_{rgb}^{i}\}_{i=1}^{N_{rgb}}$ and $\hat{X}_{ir}=\{\hat{x}_{ir}^{j}\}_{j=1}^{N_{ir}}$, where $N_{rgb}$ and $N_{ir}$ denote the numbers of visible and infrared tokens, respectively. A temperature-scaled similarity matrix $S\in\mathbb{R}^{N_{rgb}\times N_{ir}}$ is then computed as
\[
S(i,j)=\tau^{-1}\,\langle W_{rgb}\hat{x}_{rgb}^{i},\,W_{ir}\hat{x}_{ir}^{j}\rangle,
\]
where $W_{rgb}$ and $W_{ir}$ denote learnable linear projections, $\langle \cdot,\cdot\rangle$ denotes the inner product, and $\tau$ is a temperature parameter. Row- and column-wise softmax yield bidirectional confidence matrices:
\[
\begin{aligned}
	P_{rgb\rightarrow ir}(i,j)
	&=\operatorname{Softmax}_{j}\!\bigl(S(i,\cdot)\bigr),\\
	P_{ir\rightarrow rgb}(i,j)
	&=\operatorname{Softmax}_{i}\!\bigl(S(\cdot,j)\bigr).
\end{aligned}
\]

For each visible token, we select the infrared token with the highest confidence in $P_{rgb\rightarrow ir}$; conversely, for each infrared token, we select the most confident visible token in $P_{ir\rightarrow rgb}$. Candidates whose confidence is below $\theta_c$ are discarded. The resulting coarse correspondence set is
{\scriptsize
	\[
	\begin{aligned}
		\mathcal{M}_{c}
		= &\left\{(i,j)\,\middle|\, j=\arg\max_{j'}P_{rgb\rightarrow ir}(i,j'),\;
		P_{rgb\rightarrow ir}(i,j)\ge\theta_{c}\right\}\cup \\
		&\left\{(i,j)\,\middle|\, i=\arg\max_{i'}P_{ir\rightarrow rgb}(i',j),\;
		P_{ir\rightarrow rgb}(i,j)\ge\theta_{c}\right\}.
	\end{aligned}
	\]
}

Unlike strict one-to-one assignment, this bidirectional strategy naturally preserves one-to-many candidates at the coarse stage, improving robustness to large geometric discrepancies. In addition, invalid padded regions and boundary tokens are masked to suppress unreliable matches. The resulting $\mathcal{M}_{c}$ serve as initial matches for the subsequent fine-grained refinement and affine transformation estimation.

\subsubsection{Fine-Grained Matching}
Based on the coarse matches $\mathcal{M}_{c}$, fine-grained matching further refines each pair within local neighborhoods. Specifically, for each coarse correspondence $(i,j)\in\mathcal{M}_{c}$, three pairs of local windows centered at the matched locations are extracted in a coarse-to-fine manner, with sizes $1\times1$, $3\times3$, and $5\times5$, respectively. The $1\times1$ and $3\times3$ windows are first jointly encoded through self- and cross-attention, injecting coarse context into the finer neighborhood. The refined $3\times3$ features then guide the $5\times5$ windows in the same manner, progressively enhancing each coarse match from coarse contextual guidance to a more discriminative fine-scale neighborhood. Denoting the resulting local features as $\tilde{f}_{rgb}^{5\times5}$ and $\tilde{f}_{ir}^{5\times5}$, their similarity matrix $S_f^{(i,j)}$ are constructed as
\[
S_{f}^{(i,j)}(u,v)=\tau^{-1}\langle \tilde{f}_{rgb}^{5\times5}(u),\,\tilde{f}_{ir}^{5\times5}(v)\rangle,
\]
and $u,v$ index the local tokens in the two $5\times5$ windows. A dual-softmax operation is then applied to $S_{f}^{(i,j)}$ to obtain the local confidence matrix
{\small
	\[
	P_{f}^{(i,j)}(u,v)=\operatorname{Softmax}_{v}\!\big(S_{f}^{(i,j)}(u,\cdot)\big)\cdot
	\operatorname{Softmax}_{u}\!\big(S_{f}^{(i,j)}(\cdot,v)\big).
	\]
}

For each coarse correspondence, the highest-confidence local pair is retained when its confidence exceeds the threshold $\theta_{f}$, which suppresses ambiguous responses in the local neighborhood and yields a more reliable fine-grained match. To further reduce the discretization error introduced by window-based matching, the selected visible and infrared features are concatenated and fed into a lightweight regressor that predicts coordinate offsets for both modalities. The corrected local coordinates are mapped back to image space, yielding the final refined correspondence. All refined pairs form $\mathcal{M}_{f}$ for subsequent affine estimation.

\subsubsection{Feature Affine Registration}
Given the refined correspondence set $\mathcal{M}_{f}$, CMAA estimates an image-level affine transformation from the infrared modality to the visible reference. Let $\mathcal{M}_{f}^{+}=\{(\mathbf{p}_{rgb}^{n},\mathbf{p}_{ir}^{n},w_n)\}_{n=1}^{N}$ denote the confidence-filtered correspondences after pixel refinement, where $\mathbf{p}_{rgb}^{n}$ and $\mathbf{p}_{ir}^{n}$ are matched image coordinates and $w_n$ is the matching confidence. The transformation $\mathbf{T}_{ir\rightarrow rgb}\in\mathbb{R}^{2\times3}$ is obtained through confidence-weighted affine fitting:
\[
\mathbf{T}_{ir\rightarrow rgb}
=
\arg\min_{\mathbf{T}}
\sum_{n=1}^{N}
w_n
\left\|
\mathbf{p}_{rgb}^{n}
-
\mathbf{T}\bar{\mathbf{p}}_{ir}^{n}
\right\|_2^2,
\]
where $\bar{\mathbf{p}}_{ir}^{n}=[x_{ir}^{n},y_{ir}^{n},1]^{\top}$. Low-confidence matches are discarded, and the identity mapping is used when fewer than three valid correspondences are available or when the estimated transformation is geometrically unreliable.

The estimated transformation is then applied to each infrared feature level. For a feature map of size $H_l\times W_l$, we  define the image-to-level coordinate scaling matrix as
\[
\mathbf{S}_l
=
\operatorname{diag}
\left(
W/W_l,
H/H_l,
1
\right),
\]
where $H$ and $W$ are the input dimensions. Let
$\bar{\mathbf{T}}_{ir\rightarrow rgb}\in\mathbb{R}^{3\times3}$ be the homogeneous form of
$\mathbf{T}_{ir\rightarrow rgb}$. The affine transformation in the $l$-th feature coordinate system is given by $\bar{\mathbf{T}}_{ir\rightarrow rgb}^{\,l}=\mathbf{S}_{l}^{-1}\bar{\mathbf{T}}_{ir\rightarrow rgb}\mathbf{S}_{l}$. Since differentiable grid sampling follows an inverse mapping scheme, the aligned infrared feature is obtained as

\[
\tilde{F}_{ir}^{\,l}
=
\mathcal{W}
\left(
F_{ir}^{\,l},
\left[
\left(\bar{\mathbf{T}}_{ir\rightarrow rgb}^{\,l}\right)^{-1}
\right]_{1:2,:}
\right),
\]
where $\mathcal{W}(\cdot,\cdot)$ denotes bilinear feature warping. The resulting pyramid $\tilde{F}_{ir}=\{\tilde{F}_{ir}^{1/4},\tilde{F}_{ir}^{1/8},\tilde{F}_{ir}^{1/16},\tilde{F}_{ir}^{1/32}\}$ is used for subsequent cross-modal fusion and oriented detection.

\subsection{Illumination-Guided Complementary Fusion}

Given the visible feature pyramid $F_{rgb}=\{F_{rgb}^{l}\}_{l\in\mathcal{L}}$ and the aligned infrared feature pyramid $\tilde{F}_{ir}=\{\tilde{F}_{ir}^{l}\}_{l\in\mathcal{L}}$, where $\mathcal{L}=\{1/4,1/8,1/16,1/32\}$, IGCF performs illumination-adaptive fusion on geometrically aligned features. Although CMAA reduces spatial misalignment, visible features remain unreliable under poor illumination, whereas infrared features generally provide more stable responses. IGCF therefore suppresses unreliable visible responses under poor illumination while enhancing complementary infrared information.

The illumination condition is estimated directly from the visible image without additional annotation. Given the visible image $I_{rgb}\in\mathbb{R}^{H\times W\times C}$ with pixel intensities in $[0,255]$, the illumination score $\eta\in[0,1]$ is computed as the normalized mean intensity of $I_{rgb}$. A darkness-aware factor $d(\eta)$ is then obtained by clipping $(\eta_0-\eta)/\eta_0$ to $[0,1]$, where $\eta_0$ is an illumination threshold. Thus, $d(\eta)$ increases in darker scenes and approaches zero under sufficient illumination.

For each feature level $l$, IGCF first applies lightweight feature refinement to obtain the enhanced responses $E_{rgb}^{l}$ and $E_{ir}^{l}$. A spatial gate is generated from the discrepancy between the two responses:
\[
G^{l}
=
d(\eta)\cdot
\mathcal{G}^{l}
\left(
\left|
E_{rgb}^{l}-E_{ir}^{l}
\right|
\right),
\]
where $\mathcal{G}^{l}(\cdot)$ predicts a spatially adaptive gating map. The global factor $d(\eta)$ controls the illumination-dependent suppression strength, while the local discrepancy identifies inconsistent visible regions. IGCF further models shared and complementary cues through
\[
R_{cc}^{l}
=
\phi_{cc}^{l}
\left(
E_{rgb}^{l}\odot E_{ir}^{l},
\left|E_{rgb}^{l}-E_{ir}^{l}\right|
\right),
\]
where the element-wise product captures shared responses, the absolute difference represents modality-specific cues, and $\phi_{cc}^{l}(\cdot)$ adaptively integrates both. The final fused feature is
\[
F_{fus}^{l}
=
F_{rgb}^{l}
+
\tilde{F}_{ir}^{l}
+
(1-G^{l})\odot E_{rgb}^{l}
+
E_{ir}^{l}
+
R_{cc}^{l}.
\]

Invalid regions introduced by infrared feature warping are masked during infrared-related computation. In this way, IGCF preserves aligned features, regulates visible responses according to illumination and local discrepancy, and incorporates complementary infrared cues for robust detection.

\subsection{Alignment Quality-Consistency Gating}
\label{sec:aqcg}

Although CMAA aligns infrared feature before fusion, the estimated transformation $\mathbf{T}_{ir\rightarrow rgb}$ can be unreliable early in training. Strong detection supervision on imperfectly aligned features may misguide detector optimization and interfere with alignment learning. As alignment improves and becomes consistent with the detection objective, stronger detection supervision becomes beneficial. To this end, AQCG adaptively reweights the detection loss according to alignment quality and gradient consistency, suppressing unreliable supervision in the early stage and progressively promoting detector learning as alignment improves.

Let $\mathcal{L}_{align}^{t}$ and $\mathcal{L}_{det}^{t}$ denote the alignment and detection losses at iteration $t$. AQCG maintains an exponential moving average $\bar{\mathcal{L}}_{align}^{t}$ of the alignment loss with momentum coefficient $\beta$, and computes the raw alignment quality as $Q^{t}=\exp\!\left(-\bar{\mathcal{L}}_{align}^{t}/(\mathcal{L}_{ref}+\epsilon)\right)$, where $\mathcal{L}_{ref}$ is a stabilized reference loss after warm-up and $\epsilon$ is a small constant. A lower moving-average alignment loss therefore yields a higher $Q^{t}$, indicating more reliable geometric alignment. The quality score is then
remapped into a signed gating signal:
\[
Q_s^{t}
=
\tanh\!\left(\kappa(Q^{t}-\theta)\right),
\]
where $\theta$ is the quality threshold and $\kappa$ controls the transition sharpness. $Q^{t}<\theta$ suppresses detection learning, whereas $Q^{t}>\theta$ strengthens it.

To further avoid promoting detection when the two objectives conflict, AQCG measures the gradient consistency between alignment and detection with respect to the predicted affine transformation:
\[
C^{t}
=
\operatorname{cos\_sim}
\left(
\frac{\partial \mathcal{L}_{align}^{t}}{\partial \mathbf{T}_{ir\rightarrow rgb}},
\frac{\partial \mathcal{L}_{det}^{t}}{\partial \mathbf{T}_{ir\rightarrow rgb}}
\right).
\]

The gating score and detection weight are computed as
\[
s^{t}
=
(1-\lambda_{c})Q_{s}^{t}
+
\lambda_{c}C^{t},
\qquad
\alpha_{det}^{t}=2^{s^{t}},
\]
where $\lambda_{c}\in[0,1]$ balances alignment quality and gradient consistency. Since $Q_s^{t},C^{t}\in[-1,1]$, the detection weight is bounded by $\alpha_{det}^{t}\in[0.5,2.0]$. The overall objective is
\[
\mathcal{L}^{t}
=
\mathcal{L}_{align}^{t}
+
\alpha_{det}^{t}\mathcal{L}_{det}^{t}.
\]

The gating weight is detached during back-propagation and serves only as a training-time regulator. In this way, AQCG limits the influence of unreliable alignment while strengthening detection supervision when the transformation is accurate and optimization-consistent.

\begin{table*}[t]
	\centering
	\setlength{\tabcolsep}{4.2pt}
	\begin{tabular}{l c ccccc cc}
		\toprule
		\textbf{Method} & \textbf{Modality} & \textbf{Car} & \textbf{Truck} & \textbf{Bus} & \textbf{Van} & \textbf{Freight Car} & $\bm{\mathrm{\textbf{mAP}}_{50:95}}(\textbf{\%})$ & $\bm{\mathrm{\textbf{mAP}}_{50}}$(\textbf{\%}) \\
		\midrule
		RetinaNet \cite{lin2017focal} & \multirow{4}{*}{RGB} & 65.5 & 19.3 & 55.3 & 12.2 & 13 & 15.5 & 33.1 \\
		R$^3$Det \cite{yang2021r3det} & & 77.0 & 35.2 & 73.7 & 24.8 & 17.2 & 22.6 & 45.6 \\
		S$^2$ANet \cite{han2021align} & & 78.0 & 43.6 & 77.1 & 27.8 & 27.5 & 25.6 & 50.8 \\
		KFIoU \cite{yang2022kfiou} & & 75.6 & 25.1 & 61.6 & 16.5 & 19.5 & 18.6 & 39.6 \\
		\midrule
		RetinaNet \cite{lin2017focal} & \multirow{4}{*}{IR} & 85.8 & 30.8 & 51.8 & 10.5 & 16.5 & 20.4 & 39.1 \\
		R$^3$Det \cite{yang2021r3det} & & 89.8 & 39.7 & 83.2 & 24.6 & 30.7 & 32.2 & 53.6 \\
		S$^2$ANet \cite{han2021align} & & \textbf{90.2} & 55.9 & \textbf{87.2} & 33.9 & 40.8 & 36.0 & 61.6 \\
		KFIoU \cite{yang2022kfiou} & & 88.6 & 29.1 & 74.9 & 12.6 & 21.2 & 24.5 & 45.3 \\
		\midrule
		C$^2$Former + Faster R-CNN \cite{yuan2024c2former} & \multirow{6}{*}{RGB+IR}  & 89.3 & 28.4 & 45.7 & 31.1 & 21.2 & 17.5 & 43.1 \\
		C$^2$Former + S$^2$ANet \cite{yuan2024c2former} & & 89.9 & 51.6 & 83.5 & 33.7 & 38.8 & 31.4 & 59.5 \\
		DMM + Faster R-CNN \cite{Zhou2025DMM} & & 78.9 & 56.4 & 78.4 & 50.2 & 48.1 & 32.4 & 62.4 \\
		DMM + S$^2$ANet \cite{Zhou2025DMM} & & 85.7 & 58.7 & 86.6 & 51.7 & 50.7 & 35.1 & 66.7 \\
		COMO \cite{liu2026cross} & & 85.1 & 52.8 & 78.8 & 45.5 & 45.9 & 28.2 & 61.6 \\
		\textbf{JFRDet} (\textbf{ours}) & & 87.9 & \textbf{66.7} & 86.8 & \textbf{51.8} & \textbf{55.4} & \textbf{36.1} & \textbf{69.7} \\
		\bottomrule
	\end{tabular}
	\caption{Comprehensive comparative experiments on the DVMA dataset. We compared the JFRDet method with both single-modal and multispectral object detectors, all employing OBB detection heads. The best results are highlighted in bold.}
	\label{tab:comparative_experiments}
\end{table*}

\section{Experiments}

\subsection{Datasets and Evaluation Metrics}
\noindent\textbf{Datasets.} Experiments are conducted on the proposed DroneVehicle Misaligned (DVMA) benchmark, constructed from DroneVehicle~\cite{sun2022drone}. It contains 28,439 UAV-captured visible-infrared image pairs with oriented annotations for five vehicle categories: car, freight car, truck, bus, and van. DVMA introduces controlled affine perturbations to evaluate detection robustness under substantial cross-modal spatial misalignment. The training, validation, and test sets contain 17,990, 1,469, and 8,980 image pairs, respectively.

\noindent\textbf{Evaluation Metrics.} Following standard oriented object detection protocols, performance is reported using $\mathrm{mAP}_{50}$ and $\mathrm{mAP}_{50:95}$. Specifically, $\mathrm{mAP}_{50}$ reports AP at an IoU threshold of 0.50, reflecting performance under a relatively tolerant localization criterion. $\mathrm{mAP}_{50:95}$ averages AP over IoU thresholds from 0.50 to 0.95 at intervals of 0.05, providing a more comprehensive measure of localization accuracy.

\subsection{Experiment Details}

All experiments are implemented using the MMDetection and MMRotate frameworks with PyTorch 2.3.0 and CUDA 11.8 on two 24-GB RTX 3090 GPUs. For fair comparison, all visible-infrared image pairs are resized to a fixed resolution of $480 \times 384$. AdamW is used as the optimizer with an initial learning rate of $1\times10^{-4}$ and a weight decay of 0.05. All models are trained for 12 epochs. During training, the ground-truth annotations from the infrared modality are used as training labels, since the infrared images provide more complete target annotations in the original dataset.

\begin{figure}[t] 
	\centering
	\includegraphics[width=1\linewidth]{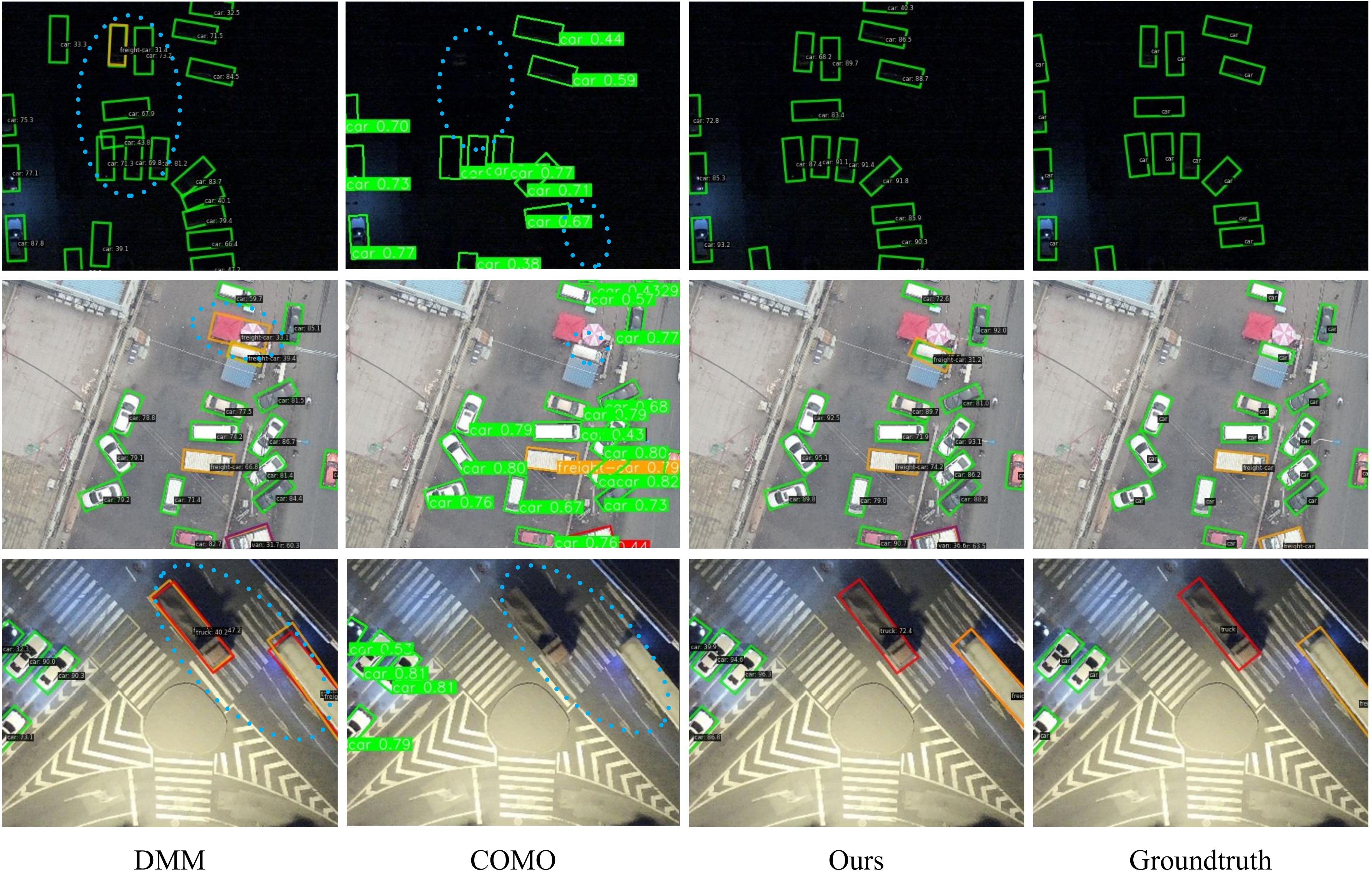} 
	
	\caption{
		Qualitative comparison on DVMA at a confidence threshold of 0.3. The last column shows visible ground truth. Blue dashed circles highlight the more accurate and complete detections of JFRDet across categories.
	}
	
	\label{vis} 
\end{figure}

\subsection{Results Comparisons}

\noindent\textbf{Quantitative comparison.} Table~\ref{tab:comparative_experiments} reports quantitative comparisons with representative single-modal and visible-infrared object detectors on DVMA. Overall, infrared detectors generally outperform their visible counterparts, indicating that infrared images provide more reliable target cues under challenging imaging conditions. S$^{2}$ANet using infrared input achieves the best single-modal performance with an $\mathrm{mAP}_{50:95}$ of 36.0\% and an $\mathrm{mAP}_{50}$ of 61.6\%, even surpassing several visible-infrared methods. This suggests that simply introducing an additional modality does not necessarily improve detection when large cross-modal geometric discrepancies exist. Although visible-infrared methods generally benefit from complementary cues, they still lack explicit geometric correction. For example, DMM with S$^{2}$ANet achieves 35.1\% $\mathrm{mAP}_{50:95}$ and 66.7\% $\mathrm{mAP}_{50}$, but remains limited under severe misalignment. In contrast, JFRDet achieves the best overall performance with an $\mathrm{mAP}_{50:95}$ of 36.1\% and an $\mathrm{mAP}_{50}$ of 69.7\%. It also obtains 66.7\% and 55.4\% $\mathrm{AP}_{50}$ on the truck and freight car categories, respectively, outperforming the best competing results of 58.7\% and 50.7\%. These gains demonstrate that explicit affine feature registration reduces geometric discrepancies and enables more reliable cross-modal fusion.

\noindent\textbf{Qualitative comparison.} Figure~\ref{vis} compares the qualitative results of DMM with S$^{2}$A-Net, COMO, and JFRDet under different illumination conditions. In low-light scenes, DMM still suffers from missed detections when RGB and infrared features are spatially misaligned, whereas JFRDet produces more complete and accurate predictions by explicitly aligning infrared features before fusion. In well-illuminated scenes, JFRDet also generates more stable oriented bounding boxes for large objects such as trucks and buses, reducing inaccurate predictions caused by unreliable cross-modal aggregation. These results demonstrate that affine feature registration effectively alleviates geometric inconsistency, while illumination-guided fusion improves the use of complementary RGB-IR cues across varying illumination conditions.

\subsection{Ablation study}

\noindent\textbf{Component-wise Ablation.} Table~\ref{tab:ablation_study} reports the component-wise ablation results on the DVMA dataset. The baseline directly fuses visible and infrared features without explicit geometric correction, achieving 66.0\% $\mathrm{mAP}_{50}$ and 34.3\% $\mathrm{mAP}_{50:95}$. After introducing CMAA, the performance increases to 67.3\% $\mathrm{mAP}_{50}$, indicating that affine alignment helps reduce cross-modal spatial inconsistency before feature fusion. Based on the aligned features, IGCF further improves the result to 68.0\% $\mathrm{mAP}_{50}$ by adaptively balancing visible and infrared cues under different illumination conditions. When AQCG is introduced, the performance reaches 68.2\% $\mathrm{mAP}_{50}$ and 35.0\% $\mathrm{mAP}_{50:95}$, showing that alignment reliability is important for stable optimization. With all components integrated, JFRDet achieves the best performance, demonstrating that CMAA, IGCF, and AQCG provide complementary benefits.

\begin{table}[t]
	\centering
	\small
	\setlength{\tabcolsep}{6.3pt}
	\begin{tabular}{cccccc}
		\toprule
		\multirow{2}{*}{\textbf{Case}}
		& \multicolumn{3}{c}{\textbf{Component}}
		& \multicolumn{2}{c}{\textbf{Performance (\%)}} \\
		\cmidrule(lr){2-4}
		\cmidrule(lr){5-6}
		& CMAA & IGCF & AQCG
		& $\mathrm{mAP}_{50}$
		& $\mathrm{mAP}_{50:95}$ \\
		\midrule
		1
		& - & - & - & 66.0 & 34.3 \\
		
		2
		& \checkmark & - & - & 67.3 & 34.4 \\
		
		3
		& \checkmark & \checkmark & - & 68.0 & 34.8 \\
		
		4
		& \checkmark & - & \checkmark & 68.2 & 35.0 \\
		
		\midrule
		\textbf{5}
		& \checkmark & \checkmark & \checkmark
		& \textbf{69.7} & \textbf{36.1} \\
		\bottomrule
	\end{tabular}
	\caption{Ablation experiments on the DVMA dataset.}
	\label{tab:ablation_study}
\end{table}

\begin{table}[t]
	\centering
	\small
	\setlength{\tabcolsep}{6.8pt}
	\begin{tabular}{c cc cc}
		\toprule
		\multirow{2}{*}{$\theta$}
		& \multicolumn{2}{c}{\textbf{w/o GC (\%)}}
		& \multicolumn{2}{c}{\textbf{w/ GC (\%)}} \\
		\cmidrule(lr){2-3}
		\cmidrule(lr){4-5}
		& $\mathrm{mAP}_{50}$
		& $\mathrm{mAP}_{50:95}$
		& $\mathrm{mAP}_{50}$
		& $\mathrm{mAP}_{50:95}$ \\
		\midrule
		0.45 & 68.8 & 35.6 & 69.4 & 36.0 \\
		0.50 & 68.6 & 36.0 & 69.7 & 36.1 \\
		0.55 & 68.6 & 35.4 & 69.2 & 36.1 \\
		0.60 & 68.7 & 35.5 & 69.0 & 35.6 \\
		\bottomrule
	\end{tabular}
	\caption{Effect of gradient consistency (GC) in AQCG.}
	\label{tab:gradient_consistency}
\end{table}

\noindent\textbf{Parameter Sensitivity of AQCG.} Table~\ref{tab:gradient_consistency} compares AQCG with and without gradient consistency, where ``w/o GC'' corresponds to $\lambda_c=0$. Introducing gradient consistency generally improves performance, indicating that measuring the agreement between alignment and detection objectives provides more reliable gating signals. Table~\ref{tab:parameter_tuning} further evaluates different quality thresholds $\theta$ and balancing factors $\lambda_c$ with gradient consistency enabled. The best result is obtained at $\theta=0.50$ and $\lambda_c=0.3$, reaching 69.7\% $\mathrm{mAP}_{50}$. Performance remains relatively stable across different settings, suggesting that AQCG is not overly sensitive to a specific hyperparameter choice. However, overemphasizing either alignment quality or gradient consistency may weaken the gating effect and lead to suboptimal performance.

\begin{table}[t]
	\centering
	\small
	\setlength{\tabcolsep}{10pt}
	\begin{tabular}{c cccc}
		\toprule
		\multirow{2}{*}{$\theta$}
		& \multicolumn{4}{c}{$\lambda_c$ / $\mathrm{mAP}_{50}$ (\%)} \\
		\cmidrule(lr){2-5}
		& 0.1 & 0.2 & 0.3 & 0.4 \\
		\midrule
		0.45 & 68.9 & 69.1 & 68.5 & 69.4 \\
		0.50 & 68.8 & 69.0 & \textbf{69.7} & 69.1 \\
		0.55 & 68.6 & 69.2 & 69.2 & 69.2 \\
		0.60 & 68.7 & 69.1 & 68.7 & 69.0 \\
		\bottomrule
	\end{tabular}
	\caption{Sensitivity analysis of AQCG with varying quality thresholds
		$\theta$ and balancing factors $\lambda_c$.}
	\label{tab:parameter_tuning}
\end{table}

\begin{figure}[t]
	\centering
	\includegraphics[width=1\linewidth]{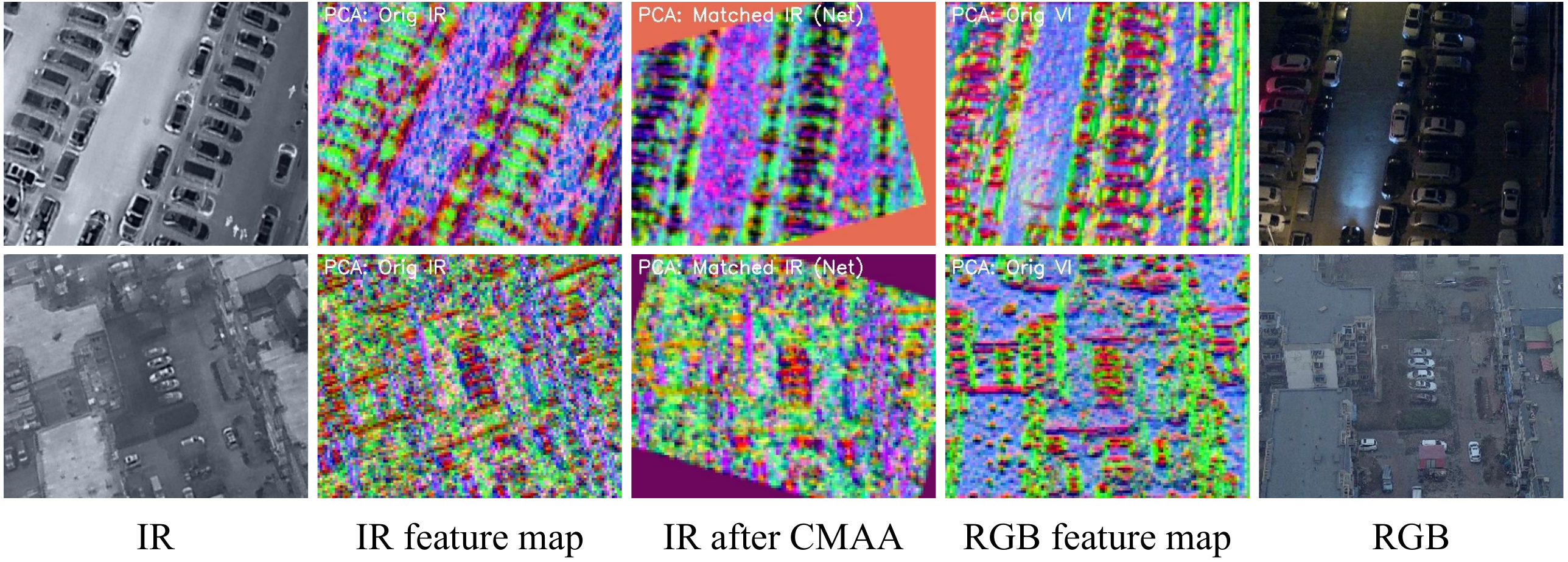}
	\caption{
		Visualization of feature alignment by CMAA. From left to right, each row shows the infrared image, the corresponding $1/4$-resolution infrared feature map, the infrared feature transformed by CMAA, the $1/4$-resolution visible feature map, and the visible image.
	}
	\label{feature_vis}
\end{figure}

\noindent\textbf{Visual Analysis of CMAA.} Figure~\ref{feature_vis} visualizes the $1/4$-resolution feature maps from the visible and infrared branches, together with the infrared features aligned by CMAA. Before alignment, the response regions of the two modalities exhibit clear spatial inconsistency, where target activations are shifted and partially mismatched. Directly fusing these features may aggregate responses from different physical locations, resulting in ambiguous cross-modal representations. After applying CMAA, the infrared responses are better registered to the visible reference. The activation centers and object structures become more consistent across modalities, while misaligned background responses are reduced. This visualization further confirms that CMAA effectively corrects feature-level geometric discrepancies and provides a more reliable basis for cross-modal fusion.

\section{Conclusion}

This paper presents JFRDet, an end-to-end visible-infrared oriented object detector for severe cross-modal geometric misalignment. Different from implicit feature adaptation under weak misalignment, JFRDet performs explicit feature-domain affine registration before cross-modal fusion, and integrates illumination-guided fusion with alignment-aware optimization. To support systematic evaluation, the DVMA benchmark is introduced to provide a challenging setting for misaligned visible-infrared oriented object detection. Experiments on DVMA show that JFRDet achieves superior performance over representative single-modal and visible-infrared detectors, while ablation studies and visual analyses further verify the effectiveness of explicit affine feature registration.


{\small
\bibliographystyle{ieee_fullname}
\bibliography{egbib}
}

\end{document}